# Generative Agent-Based Modeling: Unveiling Social System Dynamics through Coupling Mechanistic Models with Generative Artificial Intelligence

Navid Ghaffarzadegan, Aritra Majumdar, Ross Williams, Niyousha Hosseinichimeh

Department of Industrial and Systems Engineering, Virginia Tech, Falls Church, VA

Contact: navidg@vt.edu

**Abstract**

We discuss the emerging new opportunity for building feedback-rich computational models of social systems using generative artificial intelligence. Referred to as Generative Agent-Based Models (GABMs), such individual-level models utilize large language models such as ChatGPT to represent human decision-making in social settings. We provide a GABM case in which human behavior can be incorporated in simulation models by coupling a mechanistic model of human interactions with a pre-trained large language model. This is achieved by introducing a simple GABM of social norm diffusion in an organization. For educational purposes, the model is intentionally kept simple. We examine a wide range of scenarios and the sensitivity of the results to several changes in the prompt. We hope the article and the model serve as a guide for building useful diffusion models that include realistic human reasoning and decision-making.

## 1. Introduction

Through the preceding century, there was an extensive effort to employ various simulation modeling techniques to study complex systems [1-3]. Jay W. Forrester is a seminal luminary in this space, acknowledged for his pioneering contributions to modeling social systems and particularly to system dynamics [4-6] — a field that at its core embodies an endogenous approach to system structures and their evolving behaviors over time [7]. System dynamics includes a perspective wherein interactions and outcomes predominantly stem from systems' internal mechanisms rather than exogenous drivers. Put simply, the factors influencing a system's behavior are generated from within the system itself, with minimal external forces. The feedback loop is a pivotal instrument for illustrating such endogeneity; it manifests as a circular causal relationship between variables, reinforcing or balancing the system's outcome [6-8].

The system dynamics trajectory, characterized by an endogenous perspective, has endured and evolved within the various simulation modeling domains [9], including agent-based modeling, which facilitates representation of a system at the individual level [10]. When captured at that level, a system's collective outcome emerges as a consequence of individual interactions and decisions that are themselves affected by the state of the system [11, 12]. With advances in computing technologies and the increasing availability of data sources, researchers have ventured deeper into the complex modeling of multifaceted systems utilizing various statistical approaches [13]. As a result, the field of system dynamics modeling has become more interdisciplinary, requiring close cooperation among modelers, data scientists, domain experts, and stakeholders [14]. Application of such endogenous models has also expanded over the past two decades to address many pressing societal issues, from healthcare [15-17] and pandemic response [18, 19] to environmental [20, 21], organizational [22], and economic challenges [23, 24].

Representing human decision-making persists as a challenge despite these advances, regardless of whether a system dynamics model is developed using an agent-based approach or compartmental representations (stocks and flows). Models of social systems require representing humans as they obtain information and react to the state of the system. Quantifying human decision-making, though, can prove difficult [25]. Many modeling scholars, especially in economics, may typically assume humans to be rational profit-maximizers who make decisions that would benefit them the most. That approach is useful for settings with fully rational individuals or when the average of collective actions of individuals aligns with a rational profit-maximizer [26], but problems arise when dealing with conditions of systematic biases existing in human decision-making: the information is limited; decisions are myopic; people use different decision heuristics, such as rule of thumb versus empirical analysis; and satisficing (rather than maximizing) rules the preferences when it comes to comparing decision alternatives [25, 27, 28].

Behavioral decision-making scholars offer decision heuristics as a better way to represent human decision-making [28, 29]. But while the behavioral sciences may inform better representation of human behavior in dynamic models and improve model quality, the challenge nevertheless persists as there is no single, universal model of behavioral decision-making.

With advances in large language models (LLM) and the expanding scope of generative AI [30-32], social scientists now have fresh opportunities to create simulation models that more effectively represent human behavior [33, 34]. There is a growing interest in using LLMs for experimental testing, replacing or supplementing laboratory experiments for potentially harmful tests [35]. Since such models are informed and trained by extensive volumes of data, the hope is that they can lead to better representation of human behavior in computational models without the need to formulate decision rules. For example, we showed in our previous study that generative agents in a model of epidemic mimic human behaviors such as quarantining when sick and self-isolating when the number of cases increase [48]. The study also showed that a collective outcome emerges as a result of each agent responding to the state of the system, leading to multiple waves of an epidemic.

Here, we aim to contribute to this thread by articulating how to integrate systems models of societal problems with LLMs. This study provides a new way to develop behavioral system dynamics models [36]. We provide a guide to building a simple norm diffusion model at the individual level. We analyze the results for different inputs and scenarios, such as different distributions of agents' personas. This paper also aims to familiarize modelers with integrating generative AI into social systems modeling, paving the way for the development of more comprehensive models. We refer to this type of model, which couples generative artificial intelligence (AI) with an ABM, as a Generative Agent-Based Model (GABM).

## 2. Prior Studies

An LLM is a type of advanced artificial intelligence model trained on a vast amount of text data in order to understand and generate humanlike language. The main outcome of these models is coherent and contextually relevant text based on a given prompt. ChatGPT is an example of an LLM widely used by the public.

Recently, through extensive training on vast amount of web data, LLMs have demonstrated an impressive capability to generate humanlike behavior. The applications of generative agents powered by LLMs range from replacing human subjects in psychological experiments [35, 37], simulating voting patterns [38], providing support for mental wellbeing [39], exploring economic behavior [40], predicting U.S. Supreme Court decisions [41], assisting with research design and experiments [42], and encoding clinical knowledge [43]. Wang et al. [32] provides a comprehensive review of recent efforts.

While most recent studies aim at easing psychological experiments, a newly evolving opportunity exists with LLMs to study social systems by representing interacting agents in social simulations. Specifically, the investigation of multi-agent interaction powered by LLMs in simulated environments is an emerging trend with only a few, but nevertheless notable, examples. Akata et al. [44], for example, model two interacting agents in settings often studied with game theory approaches, such as the prisoner's dilemma. Moving from models of two agents with dyadic interactions to multi-agent interactions has proved to be a great opportunity to represent social contexts. Notable examples include studies conducted by Park et al. [45, 46]. They constructed a simulated environment inhabited by generative agents with predefined personas who interact and produce humanlike behavior such as comments, replies, and anti-social behaviors [45]. They then offered a sample of generated outcomes to a group of human participants who could not distinguish between the simulated behavior and actual community behavior. In a more recent application, Park et al. [46] developed a novel architecture in which generative agents could simulate believable behaviors such as waking up, cooking breakfast, going to work, and initiating conversation. This novel architecture includes a memory stream that works based on the relevance, importance, and recency of information, empowering agents to carry over some information over time.

In another application, LLM-empowered agents were used to simulate a social network [47]. These agents could observe content posted by other agents, change their attitudes and emotions, create their own content, or remain inactive. They replicated the information diffusion and change in emotion and attitude of users related to two events, the Japan Nuclear Waste Water Release Event as well as the Chained Eight-child Mother Event.

Finally, in the most recent application we cite, generative agents were harnessed to develop an epidemic model that incorporated human behavioral dynamics in response to evolving outbreaks [48]. These agents closely mirrored human actions quite impressively, adapting behaviors such as self-quarantining when unwell and isolating as cases escalated. This adaptive behavior effectively flattened the epidemic curve, generating waves of the disease [48]. This investigation underscores the potential of generative agents in capturing shifts in human behavior during epidemics.

A pivotal aspect of these models, and specifically in enhancing the representation of human decision-making aspect of them, is the creation of personas that encompass a realistic spectrum of human personalities and demographics relevant to the study context. In this approach, each individual can be described in terms of personality traits and/or demographics, which can potentially influence the LLM's responses. In some studies, personas are defined by the modelers without necessarily using a representative dataset [46], or by some demographic information inferred from data on social media [47]. The work of Williams et al. [48] is a more systematic way of defining persona in which the agent's persona is based on the commonly used Big Five traits [49] from the field of psychology. In this study, a positive versus negative version of each trait was assigned to each agent, with a chance of 50%.

The implications of these works extend beyond isolated case studies or specific systems, resonating across sectors such as healthcare and gaming. The influence of generative AI is felt throughout computational social science [34, 46, 48]. By establishing a connection between an LLM and an individual model, researchers can forge a feedback loop that intertwines individual decision-making with environmental data. This innovative approach involves the mechanistic model capturing the system's state, conveying information to the LLM, and in turn, LLM-informed decisions shaping the system's state. A notable attribute of this feedback loop is that decision rules emerge from the wealth of knowledge encapsulated in the LLM, rather than being imposed by modelers. This innovation holds the promise of refining feedback loop formulations and unlocking new avenues for system dynamics modeling in complex systems.

In this paper, we build on this evolving body of the literature and articulate the new opportunity of developing GABMs for understanding social system dynamics. What sets our research apart is its innovative integration of generative agents into the dynamic modeling landscape, with a specific focus on agent-based models. Our approach revolves around expanding the agent population within an interactive framework, tackling specific societal challenges, and systematically observing the evolution of system responses over time. We also offer a simple generic model that can be adapted for many similar problems.

## 3. What is GABM?

In a nutshell, GABM is an individual-level modeling approach in which each individual (agent) is connected to an LLM and, therefore, makes LLM-informed decisions. The GABM structure includes a cycle between the computational model built by modelers and the LLM.

To illustrate, consider a conventional agent-based models (ABMs) in which agents' decisions are formulated based on a group of predefined decision rules set by the modeler. A market diffusion model provides a good example: in such a model, agents are more likely to adopt a product if they contact an adopter, based on a preset rule. The parameter values (here the probability of adoption given contact with an adopter) is set by the modeler too.

Now imagine a model in which the modeler does not get involved in setting decision rules and related parameters. The only mechanisms modeled are the mechanics of interactions (such as the network structure of contacts). Agents' reasoning and decision-making are empowered by an LLM (in our case, ChatGPT). This requires coupling an LLM (the model of reasoning and decision) with an ABM (the model of interactions), as Figure 1 shows. In this paradigm, agents can generate reasoning and decisions. They are generative agents, and we call the coupled model a GABM.

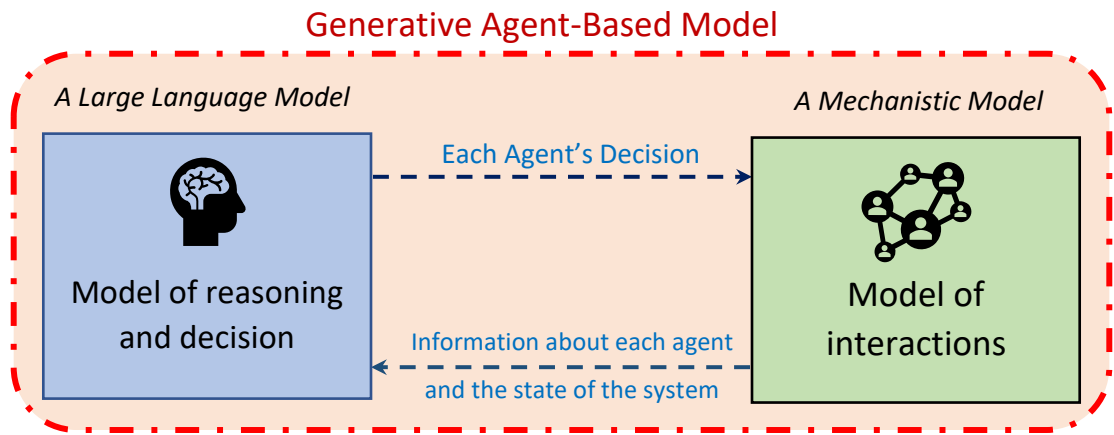

Figure 1: A conceptual diagram of a generative agent-based model coupling a model of reasoning and decision (LLM) and a mechanistic model of agents' interactions

GABM provides decision-making ability for individual agents as they interact. It includes a lot of back-and-forth between ChatGPT and the mechanistic model of agent interactions (Figure 1). For example, in Williams et al.'s [48] GABM of an epidemic, the mechanistic model informs all agents each "morning" about the prevalence of the virus (as if they listen to a daily news report or read the newspaper) and whether they have a mild cough or fever. ChatGPT then reasons and makes the decision about whether an agent goes to work or stays home. Then, based on the mechanistic model, agents that decide to leave their homes have contact with other agents and possibly transmit the disease.

In summary, the idea behind GABM is to combine the mechanistic modeling with the data-driven decision-making provided by LLM. In this paper, our case study is simpler and more general; we develop a simple GABM of diffusion dynamics.

## 4. Green or Blue? A Simple Model of Diffusion of Norms

The history of information diffusion models goes more than half a century, with the seminal work of Frank Bass [1]. At their core, information diffusion models are mathematical frameworks used to represent how new products, innovations, or ideas spread within a population over time. They have a wide range of applications from marketing and economics to sociology, management, and political sciences. Such models have been shown to be powerful tools for tracking various social phenomena such as the spread of news or norms. We focus our GABM introduction on building such a model for a specific context.

In this paper, we model a simple case of the diffusion of a norm with respect to workplace attire. Imagine 20 workers in an office who see each other every day and wear either a green shirt or a blue shirt. The workers are generative agents, which means they can reason and make decisions using an LLM. In every time step, we provide each individual with a prompt that includes information about the context, individuals' personality, choice of color for their shirt on the preceding day, and the number of people in the office who wore green or blue on the preceding day. Agents should then decide on the color they want to wear to the office that day, and for that purpose they contact an LLM.

For this model, we use ChatGPT (specifically GPT-3.5-Turbo) as our LLM. We run the model for a period of seven time-steps to investigate the possibility of generative agents converging around a norm. In contrast to conventional models, we don't set any decision rules, and we don't suggest a shirt color for the agents. In the following sections, we review the model's algorithm and explain the code.

To use the model as the tutorial there is a set of prerequisites (listed below). The Appendix includes a link where you can download the model. The codes will not run without setting an API key, which we discuss in the next sections.

### 4.1. Prerequisites

We code our model in Python and our instruction to replicate model outcomes relies on using Python. A minimum level of coding skills in Python will help quickly implement this model. However, even a user lacking a Python background should be able to proceed. If you do not have Python, you can use Google Colaboratory.[1] Click on "File" and then "new notebook" and your Google Colaboratory notebook should open. Once you open a new notebook, it will be saved in your Google Drive for access later.

To use ChatGPT, you will need an account in OpenAI.[2] Open AI offers connecting through API to ChatGPT for a relatively small fee. We estimate that running the model we describe here will cost you US$0.10 per run. To obtain your OpenAI API key, follow these steps: Step 1: Sign up for Open AI and go to Billing Overview. Click on Payment Methods and add your billing details; Step 2: Go to OpenAI Platform. Click on "Create new secret key." Provide a name for the key in the popup and click on "Create secret key." Once you create the secret key, you must copy and paste it somewhere secret. It is important to keep the API key a secret. After your OpenAI API is set, note **openai.api_key** in our code, and place your API key where it says "your-api-key-here" (insert inside the quotation).

---

[1] https://colab.research.google.com/
[2] https://openai.com/

## 4.2. Model algorithm

Figure 2 is a flowchart of the code logic and indicates the purpose of each cell of code. There are five cells of code in Python. In the first step, required Python libraries are imported (cell 1). In the second step, a "world" is created to represent the organization and workers; it is initialized with a certain number of workers and their color of choice on day zero (cell 2). In the third step, a worker class is designed to perform the process of individual decision-making by leveraging ChatGPT (cell 3). The process then continues until the final simulation round (cell 4). Finally, the results are reported (cell 5).

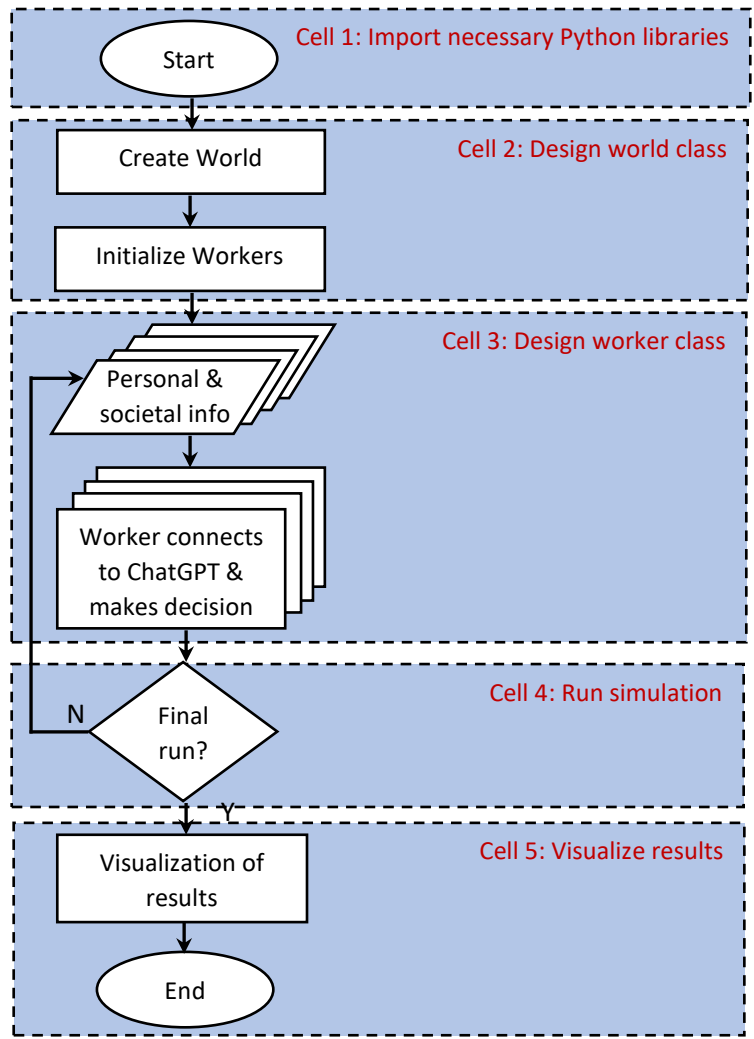

Figure 2. Flowchart of green or blue shirt model code logic

## 5. Simulation results

### 5.1. A sample of simulation runs

We expect everyone will be able to run the provided code (see the Appendix) after inserting their API keys in the specified place.[3] Once simulation is completed, running the final cell will produce a graph-over-time of the number of people who wear blue as well as a table that includes each agent's decision at each time step (1 for wearing a blue shirt and 0 for green).

A wide range of results can be obtained from the model depending on the initial condition, which is randomly set in the current version, with the equal chance of wearing blue versus green at time zero (check cell 2 the world class). Figure 3 shows two different runs that one can generate. Figure 3, Panel A, presents the results when 10 workers wear blue initially; Figure 3, Panel B shows the results created by changing the initial number of workers wearing blue to 8.  As shown, one of the two colors eventually takes over, as if a norm has evolved around wearing a specific color.

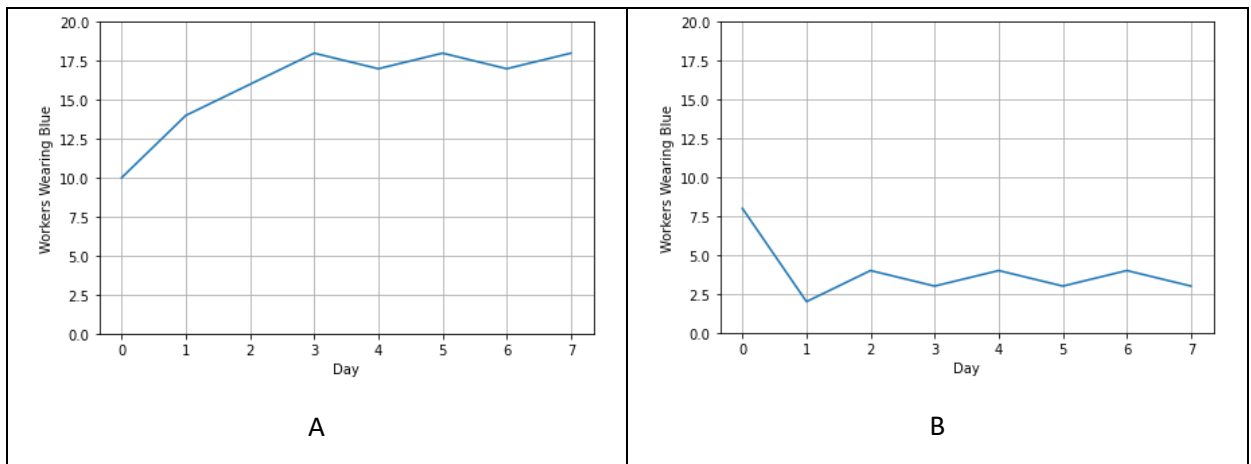

**Figure 3. Two examples of simulation run.**

Table 1 shows a sample of results corresponding to Figure 3, Panel A that were generated by the GABM model. The table can help in examining individuals' decision over time as affected by their personas, the shirt choices of the rest of the population, and individuals' own choices of shirt color the previous day. For example, while Mark switches to blue on day 1 and continues to wear blue for the remainder of the simulation, Peter decides not to follow his co-workers switching from blue to green, and Mia decides to change her shirt color every day.

---

[3] To run the model in Python, you need to run all the cells individually. To re-run the model, you should run all other cells *except* the first one again. We expect each run to take around 5 minutes. If your program is "paused and retrying," it is due to the high load on OpenAI's servers; in most cases, it will continue to run after a few minutes. You can view the status of OpenAI's servers at https://status.openai.com/.

**Table 1. Choices of each agent over time**

|  | Day 0 | Day 1 | Day 2 | Day 3 | Day 4 | Day 5 | Day 6 | Day 7 |
|---|---|---|---|---|---|---|---|---|
| **Adrian** | 1 | 1 | 1 | 1 | 1 | 1 | 1 | 1 |
| **Mark** | 0 | 1 | 1 | 1 | 1 | 1 | 1 | 1 |
| **Greg** | 0 | 1 | 1 | 1 | 1 | 1 | 1 | 1 |
| **John** | 0 | 1 | 1 | 1 | 1 | 1 | 1 | 1 |
| **Peter** | 1 | 0 | 0 | 0 | 0 | 0 | 0 | 0 |
| **Liz** | 0 | 0 | 1 | 1 | 1 | 1 | 1 | 1 |
| **Rosa** | 0 | 0 | 1 | 1 | 1 | 1 | 1 | 1 |
| **Patricia** | 0 | 1 | 1 | 1 | 1 | 1 | 1 | 1 |
| **Julia** | 0 | 0 | 0 | 1 | 1 | 1 | 1 | 1 |
| **Kathy** | 1 | 0 | 0 | 0 | 0 | 0 | 0 | 0 |
| **William** | 0 | 1 | 1 | 1 | 1 | 1 | 1 | 1 |
| **Benjamin** | 1 | 1 | 1 | 1 | 1 | 1 | 1 | 1 |
| **Mike** | 1 | 1 | 1 | 1 | 1 | 1 | 1 | 1 |
| **David** | 1 | 1 | 1 | 1 | 1 | 1 | 1 | 1 |
| **George** | 1 | 1 | 1 | 1 | 1 | 1 | 1 | 1 |
| **Emma** | 0 | 1 | 1 | 1 | 1 | 1 | 1 | 1 |
| **Olivia** | 1 | 1 | 1 | 1 | 1 | 1 | 1 | 1 |
| **Elizabeth** | 1 | 1 | 1 | 1 | 1 | 1 | 1 | 1 |
| **Isabella** | 1 | 0 | 1 | 1 | 1 | 1 | 1 | 1 |
| **Mia** | 0 | 1 | 0 | 1 | 0 | 1 | 0 | 1 |

The GABM model also provides the opportunity to study the reasoning of each individual, as this information is printed during the simulation runs. Table 2 is a sample of the reasoning agents offered for their decisions; the examples correspond to the simulation shown in Figure 3, Panel B.

**Table 2. Examples of agents' reasoning and decisions**

| |
|---|
| **Context information: Yesterday on day 0, 8 of 20 wore blue shirts.** |
| • Adrian's reasoning: Since yesterday, more people in the office wore green shirts than blue shirts, I will choose to wear a green shirt today to conform with the majority and maintain a sense of unity in the office.<br>• Adrian's response: green |
| • Julia's reasoning: Since I am a low conformist person, I would choose to wear the blue shirt today. Yesterday, more people wore green shirts than blue shirts, so wearing the blue shirt would make me stand out and be different from the majority. This could potentially help me be noticed and remembered by my superiors, increasing my chances of being successful and earning more money.<br>• Julia's response: blue |
| **Context information: Yesterday on day 3, 3 of 20 wore blue shirts.** |
| • Emma's reasoning: Since the majority of employees wore green shirts yesterday, I will choose to wear a green shirt today to conform and maintain harmony in the office.<br>• Emma's response: green |
| • Mia's reasoning: Since I am a low conformist person and I want to be successful, I will choose to wear the blue shirt today. Yesterday, the majority of employees wore green shirts, so by wearing the blue shirt, I will stand out and potentially be noticed by my superiors. This could increase my chances of being recognized for my work and potentially lead to more opportunities for success and earning more money.<br>• Mia's response: blue |

This example shows that a GABM can be used for both qualitative and quantitative analysis of norm diffusion dynamics, and in general many other social system problems, which provides the opportunity to understand the reasons behind changing trends. It also provides the opportunity to study each individual's decision-making and reasoning patterns over time, considering that people with different personalities might have different feedback loops affect their decisions. For example, while one person's decision might be driven mainly by peer pressure, another might be more attracted to a non-conforming choice.

### 5.2. A systematic investigation of the results

In the next steps, we systematically investigate simulation runs under different scenarios and examine simulation results sensitivity to change in the prompt. In each condition, we run the model 100 times and plot all the results together to explore the model's robustness and whether it always leads to a dominant color norm. To make the process operationally easier to run, a for-loop can be added to the model to run it multiple times. A link to the modified code is provided in the Appendix. Overall, 12 conditions are simulated and for each condition 100 simulation runs are conducted. The experiments are listed in Table 3.

**Table 3: A list of simulation experiments**

| Experiment | Date | Personas | Extra attractor | Temp. | Prompt Sequence | Agents names |
|---|---|---|---|---|---|---|
| **E1: Base run** | 8/13/23 | Conformity traits (CT) | None | 0 | Base | Base |
| **E2: No personality traits** | 8/14/23 | **None** | None | 0 | Base | Base |
| **E3: Extensive personality traits** | 8/14/23 | **CT & 3 less relevant** | None | 0 | Base | Base |
| **E4: Less relevant traits** | 8/14/23 | **3 less relevant traits** | None | 0 | Base | Base |
| **E5: Extra attractor** | 8/18/23 | CT | **Yes** | 0 | Base | Base |
| **E6: E4 & extra attractor** | 8/15/23 | **CT & 3 less relevant** | **Yes** | 0 | Base | Base |
| **E7: Temperature 0.25** | 8/15/23 | CT | None | **0.25** | Base | Base |
| **E8: Temperature 0.5** | 8/15/23 | CT | None | **0.5** | Base | Base |
| **E9: Change in prompt sequence 1** | 9/2/23 | CT | None | 0 | **Change** | Base |
| **E10: Change in prompt sequence 1** | 9/2/23 | CT | None | 0 | **Change** | Base |
| **E11: Iranian names** | 9/2/23 | CT | None | 0 | Base | **Farsi** |
| **E12: Base run (repeat)** | **9/2/23** | CT | None | 0 | Base | Base |

The results of this section are shown in Figure 4, panels A-L. As of September 2023, producing each panel takes about 3-4 hours, and costs US$10 for the API connection, yielding a total of 42 hours and US$120. We review each test and its results in the following sections. In addition, for experiments starting from the initial color choice probability of 0.5 (all experiments but E5 and E6), we conduct statistical analysis to systematically test the effect of initial condition on the final choice of colors. The results are reported in the Appendix, Table A1. If a path dependent behavior is observed, we systematically compare the outcomes with the base run, in the Appendix (Table A2).

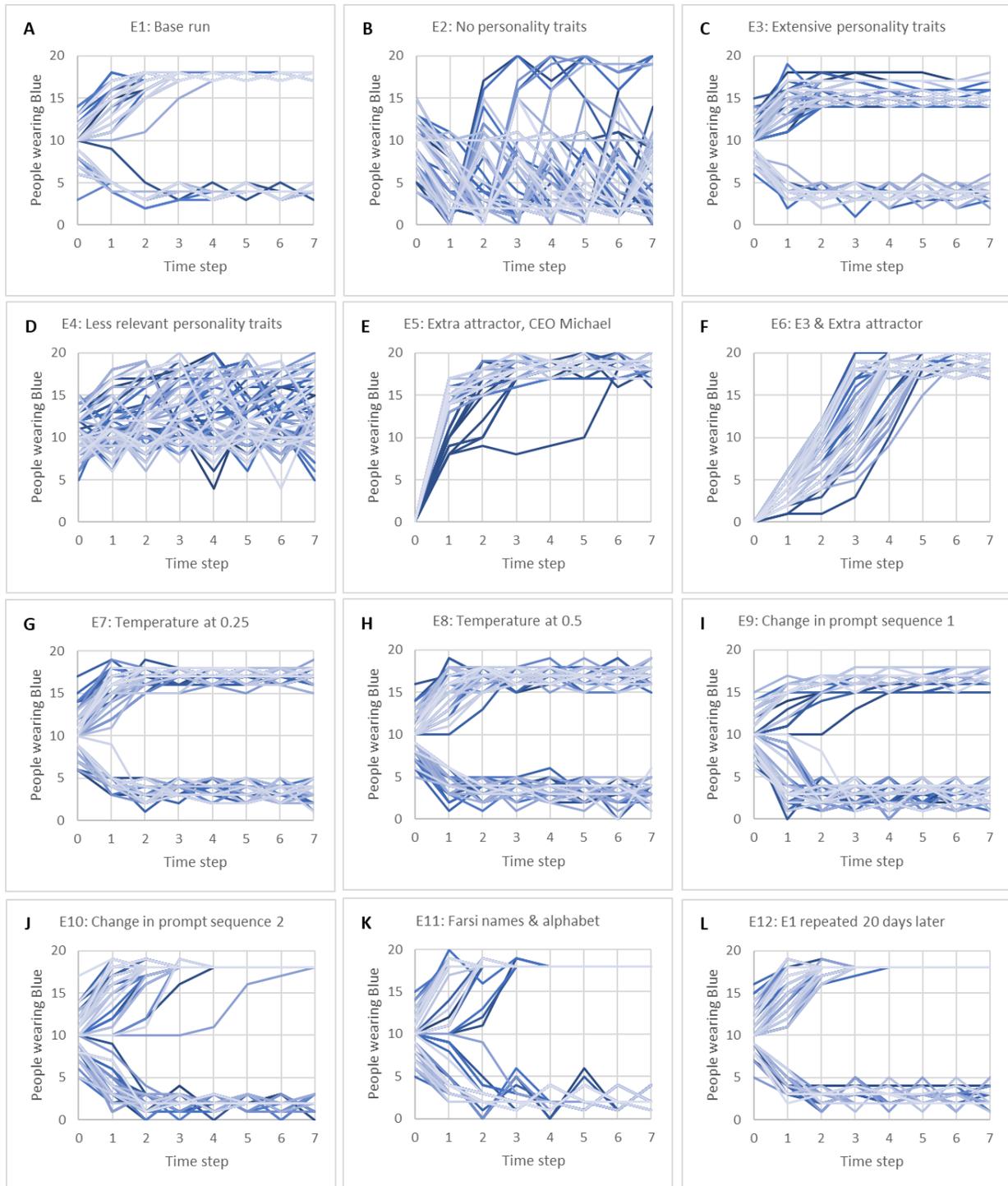

**Figure 4:** Simulation results from running the model 100 times for 12 different experiments to study the base run condition (A), and compare it with other 11 experiment to examine the effects of personas (B-D), a forceful attractor (E-F), temperature (G, H), information sequence in the prompt (I, J), agents' names (K), and simulation date for reproducing the base run (L).

### 5.2.1. Base run

Figure 4A shows the results of 100 simulations and depicts how two classes of culture emerge. In some organizations, agents collectively prefer blue while others prefer green. The choices are highly influenced by the initial condition, which in our model was random. Our statistical examination in the Appendix (Table A1, E1) confirms that the number of people wearing blue at time zero has a significant effect on the number of people wearing blue at the end of the simulation (p<.000). This path-dependent behavior seems to be a) robust, happening in all simulations, and b) does not converge to the extreme values of 0% or 100% blue colors. Specifically, the difference between the two tails is 13.3 (95% CI: 12.63 – 13.97) workers out of 20 workers (Table A1). This is due to the fact that some individuals in our model are non-conformists or low conformists, which influences their decision not to follow their peers. We note that the results are slightly biased towards blue, that is, when starting with the initial value of 10, we most often end up with the majority wearing blue (p<.000).

### 5.2.2. Effect of personas

Defined personas in GABM play an important role in the simulation. As we noted in the previous section, individuals vary in terms of their responses to the office norm. In this step, we conduct a more systematic investigation of the effect of personas. We test the effect of different personality traits used to define the agents. Considering the base run as one scenario for personas, here we conduct two other experiments: one with no personality information, and the other with more information than the base run, including some potentially less relevant personality traits to norm diffusion.

First, in order to conduct the experiment with "no personalities," we remove all personality information about the agents. Implementing this requires only removing the sentence "`You are a {self.traits} person.`" from the worker cell (cell 3) in the code. Even if the personalities are listed in cell 4, deleting this sentence will ensure that they are never used in the decision-making process. In the next experiment, we add information, including more detailed and diverse sets of traits that have little relation to conforming to social norms. To that end, we must bring back the "`You are a {self.traits} person.`" statement to the third cell, and replace the personalities in the run cell (cell 4) with those in Table 4, which adds three more dimensions per individual. The additional traits follow three of the Big Five characteristics (curious vs. cautious; friendly vs. critical; confident vs. sensitive).

**Table 4: More extensive trait information**

```
list_of_traits = ["extremely conformist, curious, friendly, and sensitive",
                  "highly conformist, cautious, friendly, and confident",
                  "conformist, curious, critical, and confident",
                  "low conformist, cautious, critical, and sensitive",
                  "non-conformist, curious, friendly, and sensitive",
                  "extremely conformist, cautious, friendly, and confident",
                  "highly conformist, curious, critical, and confident",
                  "conformist, cautious, critical, and sensitive",
                  "low conformist, curious, friendly, and sensitive",
                  "non-conformist, cautious, critical, and confident",
                  "highly conformist, curious, friendly, and confident",
                  "conformist, cautious, critical, and sensitive",
                  "conformist, curious, critical, and sensitive",
                  "conformist, cautious, friendly, and confident",
                  "low conformist, curious, critical, and confident",
                  "highly conformist, cautious, friendly, and sensitive",
```

```
                        "conformist, curious, friendly, and sensitive",
                        "conformist, cautious, friendly, and confident",
                        "conformist, curious, critical, and confident",
                        "low conformist, cautious, critical, and sensitive"]
```

Figure 4B shows that absent information about personality traits, we see no formation of dominant norms. Our statistical examination in the Appendix confirm that the number of people wearing blue at time zero has no observable effect on the number of people wearing blue at the end of the simulation (Table A1, E2). We are unsure whether these patterns are the result of the system structure or are random, but it is clear that the choice of color continues to change over time. Figure 4C shows the experiment with the comprehensive personality profiles, which demonstrates a path-dependent pattern (Table A1, E3, p<.000), although in comparison to the base run the two tails of final values are closer (Table A2, E3, p<.000). Then we tested the condition that we remove the information about "conforming" traits from personalities, resulted presented in Figure 4D. We did not note a dominant norm emerging (not shown here), concluding that the specific trait is important (Table A1, E4). Overall, as depicted, the simulation runs are sensitive to how personas are defined; therefore, it is important to define them carefully based on the real-world distribution of these traits.

### 5.2.3. Effect of a strong asymmetric adoption force

Imagine that this same organization hires a new CEO who wears a specific color that does not correspond to that worn by the majority of employees. How will generative agents react? For this test, we run two sets of experiments with different personality traits: first with the base run personalities that include only agents' conformity traits, and then with the more complete set in Table 4.

We set this test so every worker is initially wearing green and the new CEO is wearing blue. To implement this test, two parts of the model must be modified. Since we want every worker to wear green at the beginning, we need to change the initial assignment of colors. In cell 2, the world class, we initialized the workers. By modifying the line `random.random() <0.5` we can change the chance of having a blue vs. green shirt. Specifically, we change 0.5 to 0, so everyone will begin with a green shirt.

Second, we introduce the new CEO. In cell 3, the worker class, we offer prompts for agents. In that cell the quotation in red is the code thus must be changed (the prompt), despite that it looks like an informal chat. You can edit the prompt in so many different ways. Here, we want to add the information about the new CEO and the CEO's attire.

Let's add a new CEO who happens to wear blue all the time (the CEO doesn't receive feedback, or simply doesn't care). If the CEO's choice is unchanging, you can include the information in the prompt by simply adding the following sentences in the worker cell, right before "*Based on the above context, you need to choose whether to wear blue or green shirt.*" Also, fix your text to make sure the indent is consistent, that is, that the letter "M" in the word "Michael" appears exactly below the letter "O" in the word "Out of" (otherwise you will get an error).

```
Michael, the new CEO, bikes to work everyday, likes coffee, and often wears blue
shirts.
```

As we add this statement, there is one more mention of the color blue than green in the prompts. In order to balance the number of times the words appear in the prompt and avoid any possible bias, you may also add the following statement right after the information about CEO Michael:

```
You note that your neighbor who works in a different company wears green.
```

Figures 4E and 4F show the simulation results for 100 iterations. Figure 4E is for the condition in which personality traits are consistent with the base run (focus on conformity traits); Figure 4B is for the condition in which personality traits are more extensive. Both tests show the adoption of CEO Michael's style over time, with a faster adoption in the first set of experiments.

### 5.2.4. Effect of stochasticity

How random are these simulations? The parameter "temperature" in cell 3 (which has nothing to do with what you may typically think of as "temperature") controls the balance between randomness and predictability when generating text, with higher values leading to greater randomness and lower values leading to a more focused output.

We re-run the base model for temperatures equal to 0.25 and 0.5. The results in Figure 4G and 4H, along with the base run results (4A) provides a good picture of the effect of the temperature parameter in simulation runs. Overall, a dominant norm of color emerges in all conditions (Table A1, E7 and E8, $p<.000$), while the end equilibrium slightly changes in comparison to the base run (Table A2, E7 and E8).

### 5.2.5. Prompt sensitivity analysis

Prompt engineering plays an important role in using LLMs and, consequently, in GABMs. The question is how sensitive our results are to how information is presented in the prompts. On the one hand, LLMs' sensitivity to changes in prompts can be reasonable, as humans are also sensitive to how information is provided to them; but on the other hand, we prefer to know the extent to which the results are sensitive and possible explanations for such sensitivity. To that end, we conduct a set of tests changing the prompt and comparing the results with the base run.

First, we examine the robustness of the result to a change in the sequence of information provided to ChatGPT. Specifically, in one experiment, we bring the information about agents' preceding day shirt color closer to the line that asks them to decide on the color they want to wear. The results are shown in Figure 4I. In the next experiment, we move coworkers' color choices up in the prompt so that ChatGPT receives it as the first piece of information, even before agents' names and personalities. The results are shown in Figure 4J. The statistical analysis in the Appendix confirms the effect of initial condition on the final dominant choice of color (Table A1, E9 and E10, $p<.000$) although the final equilibrium values are slightly different than the base run by about 5% (Table A2, E9 and E10).

Finally, we change all agents' assigned names from ones common in the United States to common names in Iran, spelled in Farsi, to examine whether this changes the results. Figure 4K shows the results, which are qualitatively consistent with the base run. Further analysis confirms path dependency (Table A1, E11, $p<.000$) although the final equilibrium is slightly different than the base run by about 8% (Table A2, E11).

### 5.2.6. Reproducibility

Finally, we re-run the base run condition to examine how reliable ChatGPT is in reproducing the base run results. Specifically, the base run was conducted on August 13, 2013. We run the exact model that produced the base run 20 days after the initial run was conducted. Figure 4L shows the results. The path dependency pattern is clearly observable confirmed by the statistical analysis in the Appendix (Table A1, E12, p<0.000). Even though quantitively the results change slightly by not more than 8% (Table A2, E12), qualitatively they show the same patterns.

## 6. Discussions and conclusion

This study articulates the concept of GABM as a new way of modeling complex social systems. We make a case that generative AI can empower simulation models and offer a new way to develop models that endogenously represent human behavior and decision. In this individual-level approach, a mechanistic model of agents' interactions is coupled with an LLM (such as ChatGPT) and each agent makes decisions after communicating with the LLM. This innovative approach minimizes reliance on modelers' assumptions about human decision-making and instead leverages the vast data within large language models to capture human behavior and decision-making.

We illustrate the processes of developing a GABM model by showcasing a simple example involving a norm diffusion in an office, where workers must make a daily decision on their choice of shirt color between green and blue. In each time step, we provide feedback to each agent about the number of people who wore green or blue on the previous day. In contrast to conventional models, we do not set any decision rule for the agents regarding their choice of color; rather, they communicate with ChatGPT and reason and decide based on a prompt that informs them of their personality profile and provides information about the office.

A sample of simulation runs shows bifurcation based on initial conditions and an emerging office norm regarding shirt color. Specifically, agents' decisions of shirt color are influenced by their colleagues' shirt color on a previous day. While it may not be surprising that a norm emerges, as this has been shown previously in conventional models, the fact that these agents conform to an office norm without the modelers asking them to do so (i.e., there is no imposed decision model) is quite interesting. The norm emerges as they reason and make decisions after consulting with ChatGPT. In the supplementary analysis with the new CEO, it is of further interest to observe how generative agents who initially did not deviate from their peers note and adapt their choice to that of a new CEO.

This study contributes to the literature of computational social science and system dynamics modeling by offering a new method to model human decision-making that provides four distinct benefits. First, human decision-making is inherently complex, which makes it very challenging to incorporate human behavior in dynamic models. Simplifying human decision-making by assuming humans are rational does not provide reliable models for many applications. Previous research has demonstrated that the assumption of humans as purely rational decision-makers does not align with real-world decision-making behavior [28, 29]. By drawing upon the extensive dataset within an LLM, GABMs facilitate representation of human decision-making in computational models. This study thus contributes to the literature by articulating a different style of informing models about human decision-making rules.

Second, one of the criticisms of traditional agent-based modeling is that the mental model of modelers affects decision rules set in dynamic models. Agents act according to the mathematical formulations (or logic-rules) within the parameter values (e.g., probability of adoption given contact with an adopter) created by the modeler. In the GABM approach, modelers do not set any rules, and thus the model is not affected by their mental models; it appears as if the agents are autonomous. This provides several feedback loops, some of which the modeler would have not expected. For example, in our model we see that some agents decided to reject the norm, and thus an increase in one color results in the reaction of the population to a subset choosing to wear the other color.

Third, by connecting a mechanistic model to an LLM, the model uses much more data in representing human decisions than in other models. As the role of data becomes more important in parameter values and mathematical representation of the relations between the variables, GABM leverages data as a built-in feature, allowing the model to benefit from an extremely vast amount of data. Thus, the GABM approach resonates with the current emphasis on data-informed modeling, and yet uses a different approach for incorporating data on human decision-making in models.

Fourth, GABM can define different personality traits for generative agents; as a result, this new approach allows for capturing variations in decision-making in dynamic models. Modelers can create personas by indicating age, gender, personality, occupation, and other information. Different approaches have been used to define generative agents' personas, ranging from inferring demographic characteristics from data on social media [47], to use of data and relying on pretrained LLMs [45, 46], and to using the Big Five traits [49] from the field of psychology [48]. Agents' personas influence their behavior, underscoring the need for creating accurate characteristics to facilitate the precise and believable emulation of their actions. Our study shows the importance of personas in creating realistic outcomes. In this respect, our study resonates with others that suggest a more systematic ways of defining personalities [48]. When the purpose is to replicate a real-world setting, it is important to define the demographics and personality traits consistent with their real-world distributions.

This paper also provides educational contributions by illustrating how to build a GABM. The norm diffusion model presented here is intentionally kept simple for pedagogical purposes. Each step of building the model is explained, and various simulation tests are offered that help students and modelers learn by modifying the prompts. We hope the model can work as a starting point for computational social scientists to build GABMs.

Although models with generative agents address some of the limitations of prior approaches for capturing human decision-making, there are limitations that need to be considered. First, generative agents may reproduce training data when faced with recalled inquiries [35]. To address this limitation, researchers need to rephrase questions in a novel way, and dynamic modelers must define the environment with caution (e.g., to model a general pandemic, it would be necessary to use a pseudonym for the virus). Second, our results indicate that the output of the model is sensitive to some changes in the prompts. Thus, modelers need to run various sensitivity analyses. Third, generative agents are often limited by the patterns they have learned from training data and might be biased due to biases present in that training data. Modelers need to think carefully about and address such biases. In our example here, we tried to avoid this giving common names to the generative agents.

In conclusion, this article offers a novel approach, GABM, by combining the mechanistic modeling with a data-driven approach to model human decision-making powered by an LLM. The study provides a step-by-step guide to building such models using a simple example of norm diffusion in an organization. By defining human personas in GABM, such models have the potential to incorporate human behavior dynamics more realistically.

## Acknowledgments

This research is funded by US National Science Foundation, Division of Mathematical Sciences & Division of Social and Economic Sciences, Award 2229819. We are thankful for constructive comments from Kimiya MohammadiJozani, Saman Mohsenirad, Alex West, and Ann Osi.

# Appendix

**The base model**

You can download the code and run it; doing so will give you a graph and a CSV file that includes each individual's choice of color over time. To run the model in Python, you need to run all of the cells individually (note the "run" button by each cell or at the top of your screen). If you want to repeat running, you should run all but the first cell again.

Link to the base model: https://colab.research.google.com/drive/1lwPKUuoVtfnVudy9UnGiKwGeOXflOl2j?usp=sharing

**The base model with 100 iterations**

To make the process of running the code for multiple iterations automatic and efficient, we modified the code so it can run 100 times. It utilizes multiprocessing and, at the end, creates a CSV file that includes data for all the runs. Some Python IDEs (Integrated Development Environments, such as Jupyter Notebooks) may have limitations running the multiprocessing code, and so we also have a simple version that includes running the model 100 times. The code is available through (X: the multiprocessing version and X: the simple version). Note that by finding the word "iteration," you can change the number of runs from 100 to your desired number.

Link to the base model with 100 iterations: https://colab.research.google.com/drive/1PsdhOIWfB8la3PnNj-R3XeiAMD1TGFPg?usp=sharing

Link to the base model with 100 iterations and multi-processing for faster runs: https://colab.research.google.com/drive/1mbsBEN1BGwfPiePib--X5t-mqIuII_PZ?usp=sharing

**Statistical analysis**

We systematically investigate path dependency in the experiments that initially each agent started with an equal chance of wearing blue versus green (E1-E4 and E7-E12) in Table A1. The hypothesis is that slightly favoring one color at the initial point due to the stochasticity has a considerable impact on the balance of color at the end of the simulation. In Table A1, the dependent variable ($B_7$) is number of people who wear blue at time step 7, and the independent variables are two dummy variables. Noting that $\{q\} = 1$ if $q$ =True, otherwise is 0, our independent variables are $\{B_0 > 10\}$ which is 1 if more than 10 people at time 0 wear blue, otherwise zero; and $\{B_0 = 10\}$ which is 1 if an equal number of people wear blue and green at the beginning, otherwise 0.

Our regression model is:

$$B_7 = \beta_0 + \beta_1\{B_0 > 10\} + \beta_2\{B_0 = 10\}$$

We then compare models that showed a path dependency pattern (E3 and E7-E12) with the base run (E1) in Table A2. The dependent variable ($D_7$) is the absolute difference between the final point and the exact split between blue and green, that is $|B_7 - 10|$. For example, if at time 7 the number of people wearing blue is 14 or 6, then $D_7 = 4$.

Our regression model for comparing an experiment (Exp) with the base run is:

$$D_7 = \beta_0 + \beta_1\{B_0 > 10\} + \beta_2\{B_0 = 10\} + E + \beta_1\{B_0 > 10\}E + \beta_2\{B_0 = 10\}E$$

where $E$ is a binary variable which is equal to 1 if the data belongs to the experiment results that we would like to compare with the base run.

**Table A1. Association between the number of people who wear blue at the end of the simulation ($B_7$, dependent variable) and the starting point ($B_0$, independent variable) in experiments starting with a similar chance of wearing blue vs green.**

| Variable | E1 | E2 | E3 | E4 | E7 | E8 | E9 | E10 | E11 | E12 |
|---|---|---|---|---|---|---|---|---|---|---|
| $\{B_0 > 10\}$ | 13.30*** | 3.17* | 11.21*** | 0.21 | 13.88*** | 13.68*** | 13.44*** | 16.02*** | 15.65*** | 15.29*** |
| | (0.34) | (1.46) | (0.23) | | (0.36) | (0.21) | (0.67) | (0.47) | (0.71) | (0.14) |
| $\{B_0 = 10\}$ | 12.39*** | 2.33 | 11.33*** | 1.44 | 13.30*** | 13.98*** | 3.98*** | 13.74*** | 9.78*** | 15.29*** |
| | (0.41) | | (0.26) | | (0.44) | (0.26) | (0.79) | (0.66) | (0.92) | (0.16) |
| Constant | 4.36*** | 4.80*** | 4.00*** | 12.82*** | 3.46*** | 3.32*** | 3.32*** | 1.98*** | 2.35*** | 2.71*** |
| | (0.23) | (0.70) | (0.17) | (0.59) | (0.27) | (0.15) | (0.43) | (0.31) | (0.46) | (0.11) |
| R-squared | 0.95 | 0.08 | 0.97 | 0.02 | 0.94 | 0.98 | 0.80 | 0.93 | 0.84 | 0.99 |
| N | 100 | 100 | 100 | 100 | 100 | 100 | 100 | 100 | 100 | 100 |

***P<0.000; **P<0.001; *P<0.05; Different models use data from different experiments (E2-E4, E7-E12); $\{q\} = 1$ if $q$ =True, otherwise is 0.

**Table A2. Statistical comparison of experiments (E3, E7-E12) with the base run (E1).** Dependent variable is the absolute difference between number of people with blue shirt on day 7 and 10, i.e., $|B_7 - 10|$.

| Variable | E3 | E7 | E8 | E9 | E10 | E11 | E12 |
|---|---|---|---|---|---|---|---|
| $\{B_0 > 10\}$ | 2.02*** | 2.02*** | 2.02*** | 2.02*** | 2.02*** | 2.02*** | 2.02*** |
| | (0.17) | (0.16) | (0.17) | (0.20) | (0.10) | (0.16) | (0.12) |
| $\{B_0 = 10\}$ | 1.81*** | 1.81*** | 1.81*** | 1.81*** | 1.81*** | 1.81*** | 1.81*** |
| | (0.21) | (0.19) | (0.21) | (0.24) | (0.12) | (0.20) | (0.15) |
| E | 0.36* | 0.90*** | 1.04*** | 1.04*** | 2.38*** | 2.01*** | 1.65*** |
| | (0.18) | (0.16) | (0.24) | (0.19) | (0.10) | (0.15) | (0.13) |
| E* $\{B_0 > 10\}$ | -2.80*** | -1.22*** | -1.70*** | -1.94*** | -2.04*** | -1.67*** | -1.31*** |
| | (0.25) | (0.23) | (0.24) | (0.28) | (0.14) | (0.23) | (0.18) |
| E*$\{B_0 = 10\}$ | -2.47*** | -1.02*** | -1.19*** | -1.89*** | -1.83*** | -1.59*** | -1.10*** |
| | (0.3) | (0.28) | (0.29) | (0.34) | (0.18) | (0.29) | (0.21) |
| Constant | 5.64*** | 5.64*** | 5.64*** | 5.64*** | 5.64*** | 5.64*** | 5.64*** |
| | (0.12) | (0.11) | (0.12) | (0.14) | (0.07) | (0.11) | (0.08) |
| R-squared | 0.61 | 0.53 | 0.47 | 0.38 | 0.81 | 0.58 | 0.73 |
| N | 200 | 200 | 200 | 200 | 200 | 200 | 200 |

***P<0.000; **P<0.001; *P<0.05; Different models use data from different experiments (E2-E4, E7-E12); $\{q\} = 1$ if $q$ =True, otherwise is 0.

24